\documentclass[letterpaper, 10 pt, conference]{ieeeconf}  

\IEEEoverridecommandlockouts                              

\usepackage{amsmath,amsfonts}
\usepackage{amssymb}
\usepackage{bbm}
\usepackage{xcolor}
\usepackage{algorithmic}
\usepackage{algorithm}
\usepackage{booktabs}
\usepackage{array}
\usepackage[caption=false,font=normalsize,labelfont=sf,textfont=sf]{subfig}
\usepackage{textcomp}
\usepackage{stfloats}
\usepackage{url}
\usepackage{verbatim}
\usepackage{graphicx}
\usepackage{cite}
\usepackage{lipsum}  

\graphicspath{ {./figures/} }

\title{\LARGE \bf Self-Supervised Curriculum Generation for Autonomous \\ Reinforcement Learning without Task-Specific Knowledge}

\author{Sang-Hyun Lee$^{1}$ and Seung-Woo Seo$^{1}$
\thanks{This work was supported by the Institute of New Media and Communications, and the Automation and Systems at Seoul National University. (\textit{Corresponding author: Sang-Hyun Lee})}
\thanks{$^{1}$Sang-Hyun Lee and Seung-Woo Seo are with the Department of Electrical and Computer Engineering, Seoul National University, Korea        
        {\tt\small (e-mail: slee01@snu.ac.kr; sseo@snu.ac.kr)}}%
}

\begin{document}

\maketitle
\thispagestyle{empty}
\pagestyle{empty}

\begin{abstract}
A significant bottleneck in applying current reinforcement learning algorithms to real-world scenarios is the need to reset the environment between every episode. This reset process demands substantial human intervention, making it difficult for the agent to learn continuously and autonomously. Several recent works have introduced autonomous reinforcement learning (ARL) algorithms that generate curricula for jointly training reset and forward policies. While their curricula can reduce the number of required manual resets by taking into account the agent's learning progress, they rely on task-specific knowledge, such as predefined initial states or reset reward functions. In this paper, we propose a novel ARL algorithm that can generate a curriculum adaptive to the agent's learning progress without task-specific knowledge. Our curriculum empowers the agent to autonomously reset to diverse and informative initial states. To achieve this, we introduce a success discriminator that estimates the success probability from each initial state when the agent follows the forward policy. The success discriminator is trained with relabeled transitions in a self-supervised manner. Our experimental results demonstrate that our ARL algorithm can generate an adaptive curriculum and enable the agent to efficiently bootstrap to solve sparse-reward maze navigation and manipulation tasks, outperforming baselines with significantly fewer manual resets.
\end{abstract}


\section{INTRODUCTION}
Humans have a remarkable ability to continually learn and improve on their own. Reinforcement learning (RL) provides an appealing framework to empower robots with this ability. However, applying current RL algorithms to real-world scenarios presents significant challenges. One of the key challenges is the need to reset the environment after each episode \cite{levine2016end, yahya2017collective, andrychowicz2020learning, kalashnikov2021mt}. While resetting the environment is straightforward in simulated settings, it requires substantial human intervention and supervision in the real world. Furthermore, during the manual reset process, the agent cannot collect transitions through interaction with the environment, resulting in poor sample efficiency. It is clear that minimizing the manual resets required to train the agent is crucial for scaling current RL algorithms to real-world environments.

Conventional approaches to automating RL algorithms without manual resets leverage scripted reset behaviors or additional instrumentation \cite{levine2018learning, nagabandi2020deep, zhu2019dexterous, zeng2020tossingbot, sharma2020emergent}. These tailored reset mechanisms have poor scalability as they utilize heuristic rules for particular scenarios. To overcome their limitation, several recent works have introduced autonomous RL (ARL) algorithms that jointly train a reset policy to reset the environment and a forward policy to solve a task \cite{eysenbach2017leave, kim2022automating, sharma2021autonomous}. The key idea behind these ARL algorithms is to generate a curriculum that determines when to abort an episode or where to reset the agent based on the agent's learning progress. While these algorithms can reduce the manual resets required to train the agent, they depend on task-specific knowledge, such as specific initial states, reset reward functions, or demonstrations, to create their curricula. Constraining initial states to specific state space can cause the agent to fail to achieve robust performance \cite{sharma2022autonomous}. Furthermore, designing reset reward functions or demonstrations is often not straightforward and causes additional human intervention.

In this paper, we propose a new ARL algorithm that can generate a curriculum adaptive to the learning progress of the agent without task-specific knowledge. Our adaptive curriculum provides the agent with diverse and informative initial states based on its learning progress. To do so, we introduce a success discriminator and jointly train it with the reset policy: the reset policy is trained to continuously explore unseen states, and the success discriminator is trained to estimate the probability of solving a task from each initial state when the agent follows the forward policy. With these learnable models, our algorithm can identify which initial state encourages the agent to obtain diverse and informative transitions without extrinsic intervention.

The simplest approach to obtaining supervisory signals for training the success discriminator is to empirically estimate the success probabilities by collecting multiple rollouts of the forward policy for each initial state. However, this approach requires extrinsic resets between every rollout and access to uniform initial state distribution, which are both obviously impractical in the real world. To address this challenge, we relabel rollouts obtained from the forward policy and use them to train the success discriminator in a self-supervised manner. This relabeling is based on our hypothesis that we can regard every state within successful rollouts as an initial state from which the agent can solve a task using the forward policy, and vice versa. Our training procedure allows the success discriminator to adapt to continuously changing forward policy, leading to an adaptive curriculum. As the performance of the forward policy improves, the success discriminator allows the agent to reset with more diverse and informative initial states over a broader state space.

The main contribution of our work is to propose a new ARL algorithm that can generate an adaptive curriculum without task-specific knowledge. It is in contrast to previous ARL algorithms that require task-specific knowledge to generate adaptive curricula. We evaluate our algorithm against baselines on diverse maze navigation and manipulation tasks with sparse rewards. These sparse-reward tasks require the agent to do efficient exploration for long-term gains, which poses a challenge even for standard RL algorithms with manual reset access. The experimental results demonstrate that our algorithm can generate an adaptive curriculum without task-specific knowledge, enabling the agent to tackle these challenging tasks with efficient bootstrapping. Furthermore, our algorithm achieves better performance and sample efficiency than the baselines, including state-of-the-art ARL algorithms, with fewer manual resets.

\section{RELATED WORKS}
Conventional approaches to resetting environments without extrinsic intervention involve leveraging additional instrumentation or scripted reset behaviors \cite{levine2018learning, nagabandi2020deep, zhu2019dexterous, zeng2020tossingbot}. Levine et al. \cite{levine2018learning} used metal bins with sloped sides to reduce manual resets due to objects wedged into corners when collecting grasping data. Their data collection process enabled 14 robotic manipulators to gather around 800,000 grasp trials, with human intervention required only for replacing objects in the bins. Nagabandi et al. \cite{nagabandi2020deep} implemented a reset mechanism using a ramp and a pre-scripted 7-DoF Franka-Emika arm. The ramp had funnels that guided balls to a specific position, which the manipulator then picked up and returned to their initial state. Zeng et al. \cite{zeng2020tossingbot} introduced TossingBot that can pick up and throw various objects into selected boxes in the real world. Their robot was trained with minimal intervention by utilizing scripted reset behaviors that involved lifting tilted and bottomless boxes to return objects to a bin. While these works have shown the promising performance of real-world applications of RL algorithms, their tailored reset mechanisms were designed for specific scenarios, limiting the scalability of their algorithms.

ARL, which simultaneously learns both how to solve a task and how to reset the environment, has received significant attention in recent years. Eysenbach et al. \cite{eysenbach2017leave} proposed LNT that utilizes the forward and reset agents to induce a curriculum by early aborting a trial based on the reset value function, which is trained with a predefined reset reward function. Their experimental results showed that LNT solved continuous control tasks with fewer manual resets than standard RL algorithms. Kim et al. \cite{kim2022automating} introduced an extension of LNT that generates a curriculum by training the reset value function with examples of initial states. Their exampled-based ARL algorithm adopts RCE to train the reset policy and the reset value function without hand-crafted reset reward functions. While these previous works demonstrated that training the reset policy allows the agent to learn diverse tasks with fewer manual resets, they assumed that the initial state distribution is unimodal and has narrow support to prevent conflicting objectives. Restricting initial states to a narrow state space may constrain the visited states close to the initial and goal states. This can cause an agent to fail to ensure robust performance.

To provide the forward policy with a broad set of initial states, Xu et al. \cite{xu2020continual} proposed LSR that utilizes skills to discover diverse initial states. The key insight behind LSR is that the need to reset the agent with diverse initial states provides a natural setting to discover distinct skills. Zhu et al. \cite{zhu2020ingredients} introduced R3L that uses a random perturbation controller as the reset policy, ensuring that the support of the initial state distribution grows sufficiently. R3L is the most closely related to our algorithm presented in this work. However, R3L determines initial states without taking into account the learning progress of the forward policy, which can cause the agent to reset to initial states that are either too easy or too difficult. To address this problem, our ARL algorithm introduces the success discriminator to generate a curriculum adaptive to the learning progress of the forward policy. We would like to note that the success discriminator is trained in a self-supervised manner.

\begin{figure*}[t]
\begin{center}
\centerline{\includegraphics[width=0.78\textwidth]{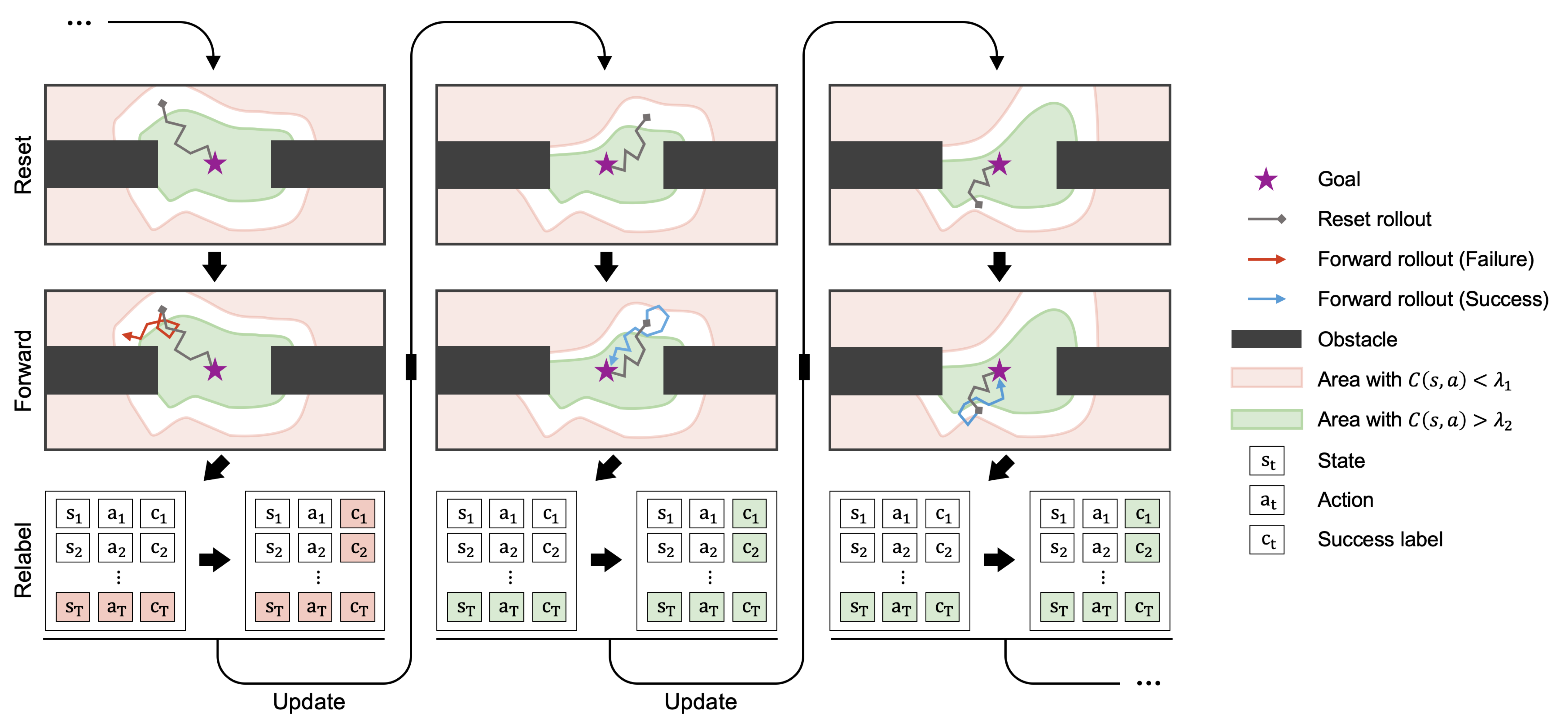}}
\vskip -0.1in
\caption{Overview of our ARL algorithm. Our algorithm generates a curriculum adaptive to the learning progress of the forward policy without task-specific knowledge. To identify the initial states that enable the agent to acquire diverse and informative transitions, we introduce the success discriminator $C(s, a)$ trained with relabeled transitions in a self-supervised manner. The subset of initial states with an estimated success probability below $\lambda_1$ is represented as the red-shaded area, while the subset of initial states with an estimated success probability over  $\lambda_2$ is represented as the green-shaded area. The goal is represented as the purple star. The rollouts from the forward policy are colored to indicate whether the agent reaches the goal or not.}
\label{fig:overview}
\end{center}
\vskip -0.23in
\end{figure*}

\section{AUTONOMOUS REINFORCEMENT LEARNING VIA SELF-SUPERVISED CURRICULUM}
Here we introduce our ARL algorithm that can generate an adaptive curriculum without task-specific knowledge. Our algorithm aims to enable the agent to continuously learn and improve on its own. Figure \ref{fig:overview} presents an overview of our algorithm. The training procedure with our adaptive curriculum consists of three phases: 1) Resetting the environment with diverse and informative initial states, 2) solving a task from initial states, and 3) updating learnable models with relabeled transitions. In the remainder of this section, we describe how our algorithm manages each phase in detail.

\subsection{Problem Formulation}
We model an environment using a Markov decision process (MDP), defined as the tuple $(S, A, P, R, \rho_0, \gamma, T)$. $S$ represents the set of states, $A$ represents the set of actions, and $P: S \times A \times S \rightarrow [0,1]$ represents the state transition model. The function $R: S \times A \times S \rightarrow \mathbb{R}$ is the reward function that outputs a scalar feedback called a reward, $r$. $\rho_0: S \rightarrow [0,1]$ is the initial state distribution, $\gamma$ is the discount factor, and $T$ is the time horizon. In this MDP, the agent selects actions based on a forward policy denoted as $\pi_f: S \times A \rightarrow [0,1]$, which maps states to a probability distribution over actions. The goal of RL is to find the optimal policy $\pi^*_f$ that maximizes the expected cumulative rewards when the state transition model is unknown.

To formulate the problem of minimizing extrinsic intervention, we introduce several distinct settings that differ from standard RL algorithms \cite{lillicrap2015continuous, mnih2016asynchronous, haarnoja2018soft}. First, while standard RL algorithms assume that the agent has access to extrinsic resets after every episode, we define the reset policy $\pi_r(a|s)$ and train it to reset the agent to initial states for subsequent episodes. Note that initial states guide the transitions the agent encounters during episodes. Second, standard RL algorithms assume that the agent has access to sophisticated reward functions, which may not be easily defined in complex tasks. In contrast, we use a sparse indicator as a reward function, denoted as $r(s,a,s') = 1(s' \in S_G)$, where it equals 1 when the agent achieves a goal and 0 otherwise. Finally, we introduce an additional learnable model called the success discriminator $C(s,a)$. This model is designed to allow our curriculum to adapt to the learning progress of the forward policy. The success discriminator takes a state-action pair as input and is trained to estimate the probability of solving a task when the agent follows the forward policy.

\vskip 0.15in
\subsection{Training Forward and Reset Policies for Minimizing \\ \text{} \text{} \text{} \text{} Extrinsic Intervention}
Our algorithm alternates between activating the reset policy to reset the agent to an initial state and the forward policy to solve a task from the initial state. 
A critical challenge in minimizing extrinsic intervention during this training procedure is to ensure that the reset policy determines initial states that are neither too challenging nor too easy for the forward policy being trained. If the initial states are too challenging, the forward policy may fail to solve the task even with sufficient time, leading to unnecessary extrinsic intervention. Conversely, if the initial states are too easy, the forward policy may struggle to gather useful information from rollouts, resulting in poor sample efficiency and suboptimal performance.

Our key idea to address the above challenge is utilizing the success discriminator proposed in this work to identify the initial states where the forward policy can not only solve a task without causing extrinsic intervention but also obtain informative transitions. To implement this idea, we activate the reset policy until the agent reaches the subset of states $S^* \in S$ as follows:
\begin{equation} \label{eq:initial_state_condition}
S^* \triangleq \{s \in S \: | \: \lambda_1 \leq C(s,a) \leq \lambda_2  \:\: \text{for} \:\: a \sim \pi_r(a|s)\}  \,
\end{equation}
where $\lambda_1$ and $\lambda_2$ are the minimum and maximum success probabilities that initial states should have to be allowed by our algorithm. This can prevent the agent from resetting with overly dangerous or non-informative initial states. 

To continuously discover novel initial states belonging to the subset $S^*$, we use an off-the-shelf exploration algorithm to train the reset policy in equation \ref{eq:initial_state_condition}. While our algorithm is compatible with any exploration algorithm, we use the random network distillation (RND) \cite{burda2018exploration} to train the reset policy as it is scalable and easy to implement. RND defines a target network $f(s): S \rightarrow \mathbb{R}^k$ that is randomly initialized and then fixed, and a trainable predictor network $\hat{f}_\theta(s): S \rightarrow \mathbb{R}^k$. RND updates the predictor network to minimize the expected prediction error $\|\hat{f}_\theta(s)-f(s)\|^2$. The prediction error can be interpreted as the novelty of a state $s$. This is based on the observation that networks typically exhibit higher prediction errors on unseen data. We use the prediction error as an intrinsic reward $\hat{r}(s,a,s') = \|\hat{f}_\theta(s')-f(s')\|^2$ to encourage our reset policy to explore novel states. Consequently, our success discriminator and reset policy can reset the agent with diverse and informative initial states in a continuous and autonomous manner. Note that any RL algorithm can be used to train the forward policy of our algorithm.

\begin{figure}[t]
\vspace{-2.0mm}
\begin{algorithm}[H]
   \caption{Overall Training Procedure of ARL via \\ Self-supervised Curriculum Learning}
   \label{alg:training}
\algsetup{linenosize=\small}
\begin{algorithmic}[1]
   \STATE Initialize reset and forward policies $\pi_r(a|s), \pi_f(a|s)$
   \vspace{0.05mm}
   \STATE Initialize reset and forward buffers $D_r, D_f$
   \vspace{0.05mm}
   \STATE Initialize target and predictor networks $f(s), \hat{f}_\theta(s)$
   \vspace{0.05mm}   
   \STATE Initialize success discriminator $C(s,a)$
   \vspace{0.05mm}   
   \FOR{each iteration}
   \vspace{0.05mm}
   \FOR{$t \leftarrow 1 \dots T_\text{reset}$}
   \vspace{0.05mm}
   \STATE Select reset action $a_t \sim \pi_r(a_t|s_t)$
   \vspace{1.00mm}
   \IF{$\lambda_1 \leq C(s_t, a_t) \leq \lambda_2$}
   \vspace{0.05mm}
   \STATE Abort and switch to forward policy
   \vspace{0.05mm}
   \ENDIF
   \vspace{0.05mm}
   \STATE Obtain reset transition $(s_t, a_t, r_t, s_{t+1})$
   \vspace{0.05mm}
   \STATE Compute intrinsic reset reward \\
          $\hat{r}_t(s_t,a_t,s_{t+1}) = \|\hat{f}_\theta(s_{t+1})-f(s_{t+1})\|^2$
   \vspace{0.05mm}
   \STATE Add transition to reset buffer \\
          $D_r \leftarrow D_r \cup \{(s_t, a_t, \hat{r}_t, s_{t+1})\}$
   \vspace{0.15mm}
   \STATE Sample batch $B_r$ from reset buffer $D_r$
   \vspace{0.05mm}
   \STATE Update reset policy $\pi_r(a|s)$ and predictor $\hat{f}_\theta(s)$
   \ENDFOR
   \vspace{0.05mm}

   \FOR{$t \leftarrow 1 \dots T_\text{forward}$}
   \vspace{0.05mm}
   \STATE Select forward action $a_t \sim \pi_f(a_t|s_t)$
   \vspace{0.05mm}
   \STATE Obtain forward transition $(s_t, a_t, r_t, s_{t+1}, c_t)$
   \vspace{0.05mm}
   \STATE Add transition to forward buffer \\
          $D_f \leftarrow D_f \cup \{(s_t, a_t, r_t, s_{t+1}, c_t)\}$
   \vspace{0.15mm}
   \STATE Sample batch $B_f$ from forward buffer $D_f$
   \vspace{0.05mm}
   \STATE Update forward policy $\pi_f(a|s)$
   \STATE Relabel success label $c_{1:T-1}$ with $c_T$ in batch $B_f$ \\
   \vspace{0.05mm}   
   \STATE Update success discriminator $C(s,a)$    
   \ENDFOR
   \vspace{0.05mm}
   \vspace{0.5mm}
   \IF{$s_{T_{\text{forward}}} \notin \mathcal{G}$}
   \vspace{0.05mm}
   \STATE Reset environment with extrinsic intervention
   \vspace{0.05mm}
   \ENDIF
   \vspace{0.05mm}
   \ENDFOR 
\end{algorithmic}
\end{algorithm}
\vspace{-5.0mm}
\end{figure}

\subsection{Training Success Discriminator for Adaptive \\ \text{} \text{} \text{} \text{} Curriculum Generation}
\label{subsection:training_success_discriminator}
The success discriminator is the key component that enables our curriculum to adapt to the learning progress of the forward policy. The straightforward approach to obtaining supervisory signals for training the success discriminator is to empirically estimate the success probabilities by collecting multiple rollouts from each initial state. Each rollout consists of sequential transitions $(s_t,a_t,r_t,s_{t+1},c_t)$, where $c_t \in \{0,1\}$ indicates whether the agent solves a task at this transition. Note that standard RL algorithms use this indicator, which we refer to as the success label in our work, to update the state-action value function $Q^{\pi_f}(s,a)$. Unfortunately, the empirical estimation approach is practically infeasible in the real world, as it requires access to a uniform initial state distribution over valid states and involves repetitive manual resets between rollouts from each initial state. 

To address this challenge, we relabel the success labels of all transitions $c_{1:T}$ with the success label of the last transition $c_T$ for each rollout, and then utilize these relabeled transitions as supervisory signals to train the success discriminator in a self-supervised manner. This relabeling strategy is based on our hypothesis that we can interpret all states within successful rollouts as initial states from which the forward policy can solve a task, and vice versa. In other words, a relabeled success label $c_t$ indicates whether the agent can solve a task in a trial when the agent takes an action $a_t$ in a state $s_t$ and follows the forward policy. The objective of our success discriminator can then be written as follows:
\begin{equation} \label{eq:discriminator_objective}
\begin{aligned}
&
\! \! \min_C - \mathbb{E}_{(s_t,a_t,c_t) \sim D_f} [c_t \log(C(s_t,a_t)) \\ 
& \: \: \: \qquad \qquad \qquad \qquad {}+ (1-c_t)\log(1-C(s_t,a_t))] \,
\end{aligned}
\end{equation}
where $D_f$ is the forward buffer and the success label $c_t$ is 1 when a trial is successful and 0 otherwise.

As the performance of the forward policy improves, the success discriminator allows the reset policy to discover diverse and informative initial states in a broader state space. Interestingly, this, in turn, leads to the performance improvement of the forward policy. In conclusion, our ARL algorithm can create a curriculum adaptive to the learning progress of the forward policy, allowing the agent to obtain diverse and informative transitions. This adaptive curriculum enables the forward policy to efficiently bootstrap on the earlier success of easier tasks to learn harder tasks. Furthermore, we would like to emphasize that, unlike previous ARL algorithms, our algorithm can generate the adaptive curriculum without task-specific knowledge such as demonstrations or sophisticated reward functions. 
Algorithm \ref{alg:training} describes the overall training procedure of our algorithm.

\section{EXPERIMENTS}
Our experiments are designed to answer the following questions: 1) Can our algorithm achieve more robust performance and better sample efficiency than state-of-the-art ARL algorithms?, 2) Can our algorithm reduce the number of manual resets required to solve a task?, 3) Can our algorithm generate a curriculum adaptive to the learning progress of the forward policy?, and 4) How does the success probability range of the initial states being sampled affect the performance of our algorithm? To answer these questions, we evaluate our algorithm against baselines on several maze navigation and manipulation tasks with sparse rewards.

\begin{figure*}[t]
\vspace{0.13cm}
\begin{center}
\centerline{\includegraphics[width=0.83\textwidth]{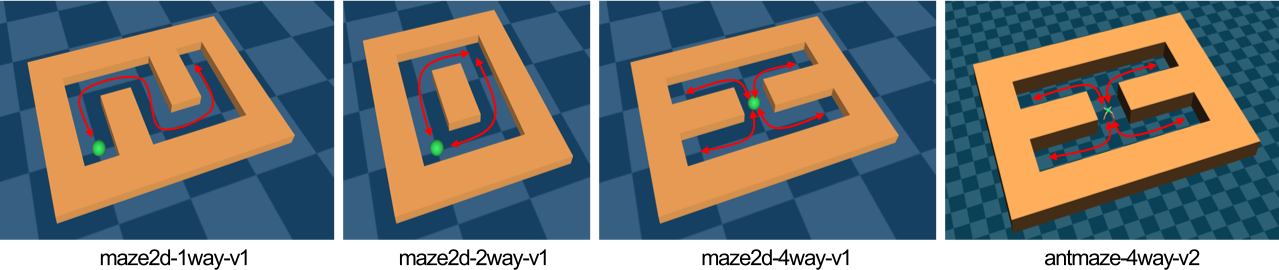}}
\caption{Maze navigation tasks introduced in our work. The locations of the goals correspond to the positions of the agents in these snapshots. The routes for each goal are represented as red lines. These tasks require the agent to reach the goals from diverse initial states without access to extrinsic reset.}
\label{fig:environments}
\end{center}
\vskip -0.18in
\end{figure*}

\begin{figure}[t]
\begin{center}
\centerline{\includegraphics[width=0.68\columnwidth]{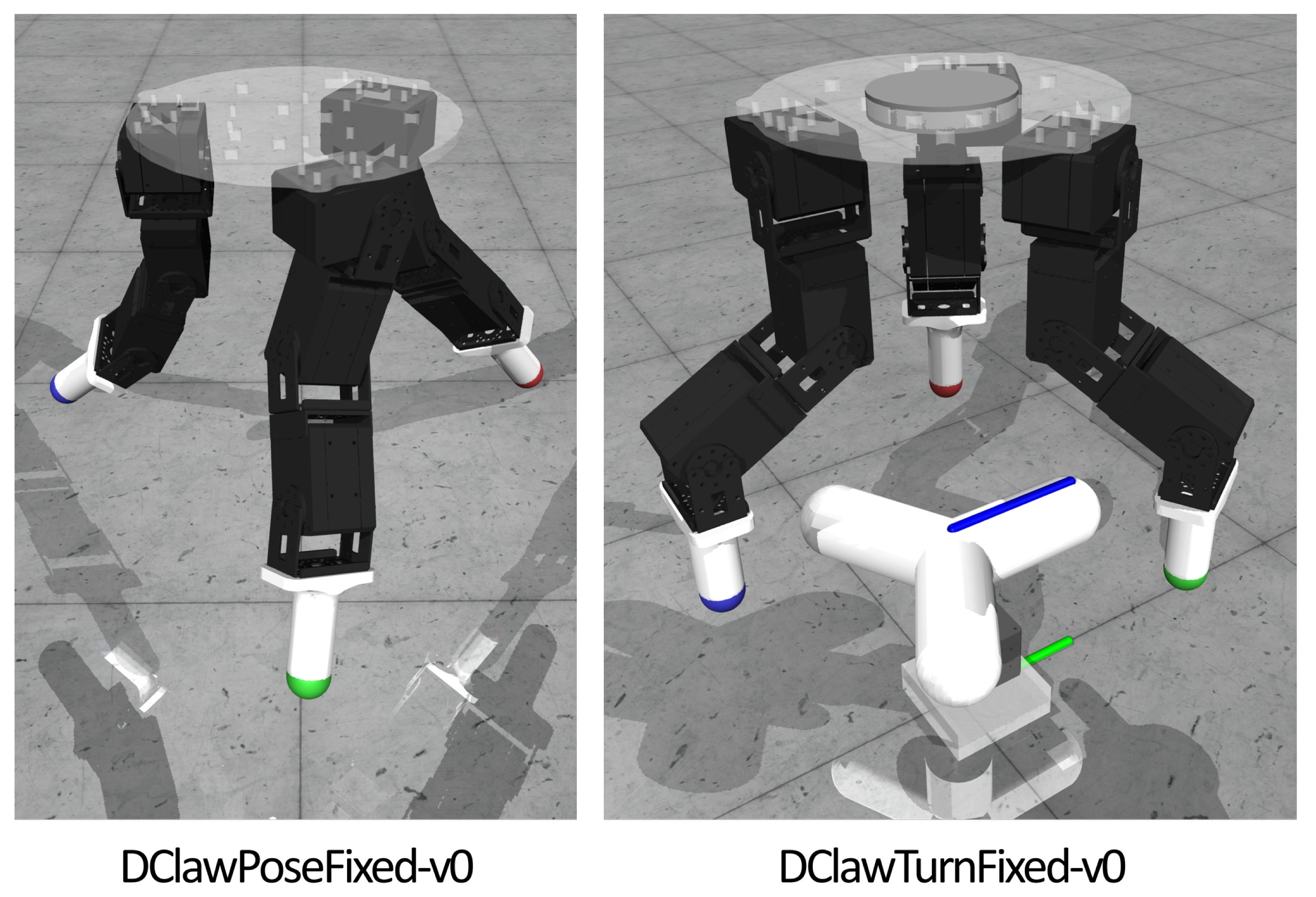}}
\caption{Manipulation tasks used in our work. These snapshots represent the target pose of a three-fingered hand robot and the target orientation of a three-pronged valve in each task.}
\label{fig:environments_dclaw}
\end{center}
\vskip -0.28in
\end{figure}

\subsection{Baselines}
The baselines used in our experiments are as follows: 1) an RL agent that has access to manual reset with specific initial states (Reset RL), 2) an RL agent that has access to manual reset with uniform initial states (Oracle RL), and two state-of-the-art ARL algorithms, 3) LNT \cite{eysenbach2017leave}, and 4) R3L \cite{zhu2020ingredients}. LNT requires a predefined reset reward function and a unimodal initial state distribution with narrow support to generate a curriculum. In contrast, R3L and our algorithm do not assume these requirements and provide the agent with diverse initial states. The main difference between R3L and our algorithm is whether or not the learning progress of the forward policy is considered: while R3L activates the reset policy to detect diverse initial states without considering the learning progress of the forward policy, our algorithm activates the reset policy and the success discriminator to detect diverse and informative initial states based on the learning progress of the forward policy. To focus on evaluating curriculum efficiency, we did not implement state embedding using VAE in R3L. Oracle RL has an impractical assumption that the agent has access to uniform initial state distribution over the valid states, but it can be interpreted as an upper bound on the performance of our algorithm.

\subsection{Environments}
Figure \ref{fig:environments} illustrates four navigation tasks used in our experiments: maze2d-1way-v1, maze2d-2way-v1, maze2d-4way-v1, and antmaze-4way-v2. All these navigation tasks are provided by D4RL \cite{fu2020d4rl}. Although these tasks may seem straightforward, they pose two significant challenges even for standard RL algorithms with access to manual resets. First, the forward reward function for these tasks is designed to output a sparse reward, with a value of 1 only when the agent reaches a goal and 0 otherwise. This requires the agent to do efficient exploration for long-term gains and bootstrapping to solve these tasks. Second, these tasks, except for maze2d-1way-v1, have multiple paths to their goals from valid states. This makes it challenging for the agent to achieve robust performance on these tasks, which requires the agent to solve them from diverse initial states. We expect that Reset RL and LNT will struggle with this challenge as they assume predefined initial states are restricted. Note that, for these baselines, we used one of the farthest states from a goal as the initial state for each task.

Figure \ref{fig:environments_dclaw} describes two dexterous manipulation tasks used in our experiments: DClawPoseFixed-v0 and DClawTurnFixed-v0. These two tasks are provided by ROBEL \cite{ahn2020robel}. The goal of DClawPoseFixed-v0 is to reposition a three-fingered hand robot to a target pose from randomly initialized poses, and the goal of DClawTurnFixed-v0 is to turn an unactuated valve from randomly initialized orientations to a target orientation by using the hand robot. DClawTurnFixed-v0 requires more structured behaviors than DClawPoseFixed-v0 because turning the unactuated valve demands a sequence of actions. Similar to the maze navigation tasks, we used a sparse indicator as the forward reward function for these tasks. 

\begin{table}[t]
\vspace{-1.0mm}
\caption{Hyperparameters}
\vspace{-3.0mm}
\begin{center}
\resizebox{0.91\columnwidth}{!}{
\begin{tabular}{c c}
\toprule
\qquad HYPERPARAMETER \qquad & \qquad VALUE \qquad \\
\midrule
\qquad Maximum Episode Step (Maze2d) \qquad & \qquad 500 \qquad \\
\qquad Maximum Episode Step (Antmaze) \qquad & \qquad 2000 \qquad \\
\qquad Batch Size \qquad & \qquad 256 \qquad \\
\qquad Replay Buffer Size \qquad & \qquad 5000000 \qquad \\
\qquad Discount Factor \qquad & \qquad 0.99 \qquad \\
\qquad Adam $\beta_1$ \qquad & \qquad 0.9 \qquad \\
\qquad Adam $\beta_2$ \qquad & \qquad 0.999 \qquad \\
\qquad Learning Rate (Reset) \qquad & \qquad 0.00003 \qquad \\
\qquad Learning Rate (Others) \qquad & \qquad 0.0001 \qquad \\
\qquad Temperature \qquad & \qquad 0.4 \qquad \\
\qquad Gradient Step \qquad & \qquad 1 \qquad \\
\qquad Target Update Interval \qquad & \qquad 1 \qquad \\
\qquad Target Smoothing Coefficient \qquad & \qquad 0.005 \qquad \\
\qquad Success Probability Range $\lambda_1 / \lambda_2$ \qquad & \qquad 0.3 / 0.7 \qquad \\ 
\bottomrule
\end{tabular}
}
\label{tab:hyperparameters}
\end{center}
\vskip -0.18in
\end{table}

\begin{figure*}[t]
\vspace{0.1cm}
\begin{center}
\centerline{\includegraphics[width=0.95\textwidth]{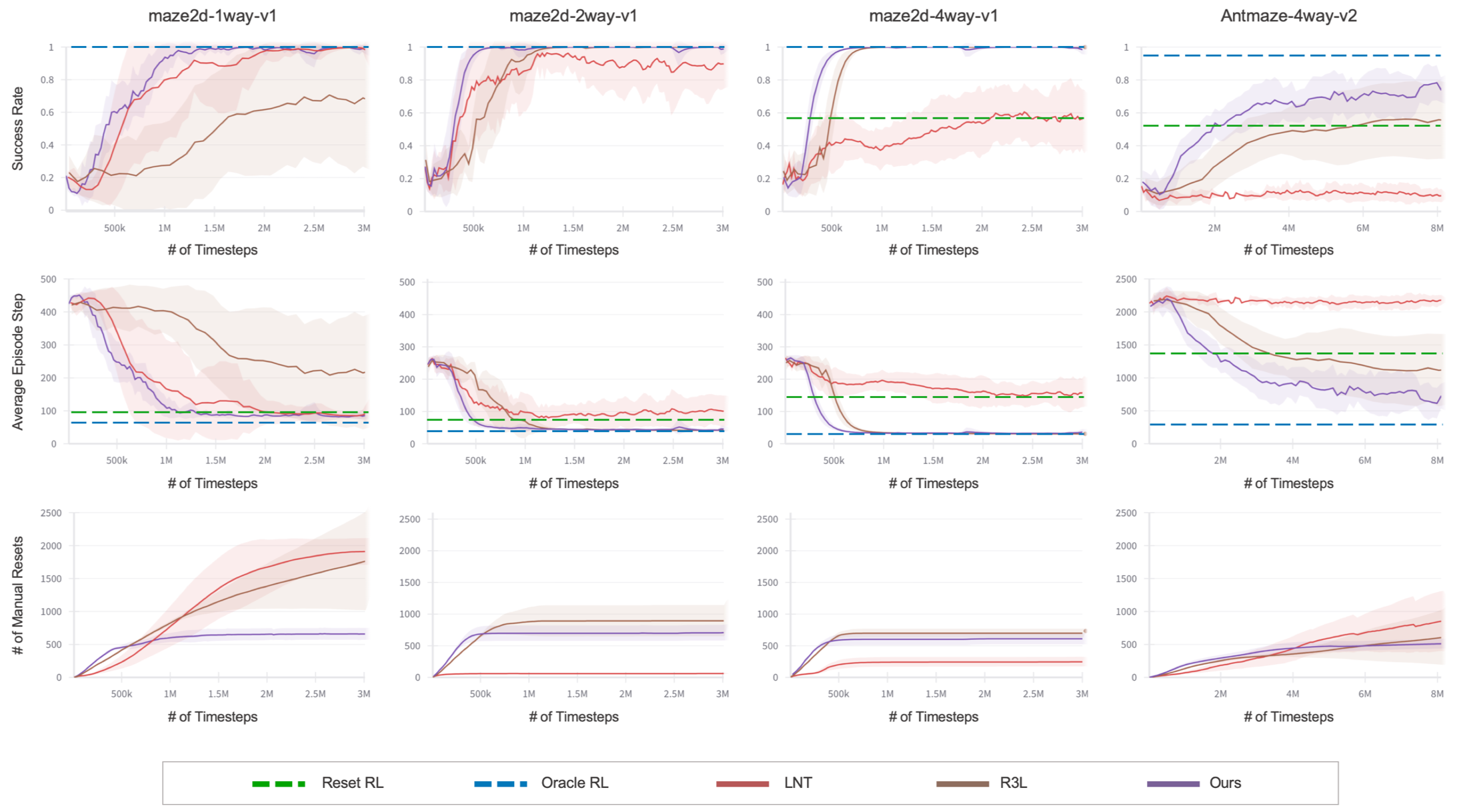}}
\caption{Learning curves for maze navigation tasks. The x-axis represents the number of training steps and the y-axis represents one of the metrics used in our experiments. The darker-colored lines and shaded areas denote the means and standard deviations over 10 random seeds, respectively. These results imply that our algorithm consistently achieves more robust performance and better sample efficiency than state-of-the-art ARL algorithms on all tasks.}
\label{fig:quantitative_results_navigation}
\end{center}
\vskip -0.1in
\end{figure*}

\begin{table*}[t]
\caption{Quantitative Results on Maze Navigation Tasks \label{tab:quantitative_results}}
\vspace{-3.0mm}
\begin{center}
\resizebox{0.9\textwidth}{!}{
\begin{tabular}{c ccc | ccc | ccc | ccc}
\toprule
& \multicolumn{3}{c}{MAZE2D-1WAY-V1} & \multicolumn{3}{c}{MAZE2D-2WAY-V1} & \multicolumn{3}{c}{MAZE2D-4WAY-V1} & \multicolumn{3}{c}{ANTMAZE-4WAY-V2}\\
\toprule
& \: AS $\downarrow$ & SR $\uparrow$ & MR $\downarrow$ \qquad 
& \: AS $\downarrow$ & SR $\uparrow$ & MR $\downarrow$ \qquad 
& \: AS $\downarrow$ & SR $\uparrow$ & MR $\downarrow$ \qquad 
& \: AS $\downarrow$ & SR $\uparrow$ & MR $\downarrow$\\
\midrule
ResetRL \qquad & 113.3 & 0.98 & 6000.0  \qquad & \: 71.5 & 0.99 & 10000.0 \qquad &  175.9 & 0.53 & 10000.0 \qquad &  1417.7 & 0.48 & 5000.0 \\
OracleRL \qquad & \: 81.5 & 1.00 & 6000.0  \qquad & \: 41.2 & 1.00 & 10000.0 \qquad & \: 30.9 & 1.00 & 10000.0 \qquad &  \: 198.3 & 0.95 & 5000.0 \\
LNT \qquad & \: 86.6 & 0.99 & 1887.6  \qquad & \: 80.1 & 0.97 & \: \: \: 63.5 \qquad & 180.0 & 0.52 & \: \: 195.8 \qquad & 2032.8 & 0.17 & \: 976.9 \\
R3L \qquad & 121.9 & 0.91 & 1410.6  \qquad & \: 48.9 & 0.98 & \: \: 810.3 \qquad & \: 31.8 & 1.00 & \: \: 715.7 \qquad &  \: 801.5 & 0.72 & \: 541.7 \\
\textbf{OURS} \qquad & \: \textbf{82.4} & \textbf{0.99} & \: \textbf{593.0}  \qquad & \: \textbf{46.4} & \textbf{0.98} & \: \: \textbf{642.7} \qquad & \: \textbf{32.1} & \textbf{1.00} & \: \: \textbf{605.7} \qquad & \: \textbf{368.0} & \textbf{0.89} & \: \textbf{499.1}\\
\bottomrule
\end{tabular}
}
\end{center}
\end{table*}

\subsection{Implementation Details}
The key learnable models in our ARL algorithm are the forward policy, the reset policy, and the success discriminator. All these models are implemented with neural networks having two hidden layers of 512 units with ReLU activations. The forward and the reset policies output the parameters of Gaussian distribution over continuous actions. The success discriminator has an additional sigmoid layer to output the probability of solving a task. We utilized Soft Actor-Critic (SAC) \cite{haarnoja2018soft}, which is a state-of-the-art off-policy RL algorithm, to train the policies. We use the Adam optimizer to update the policies and the RMSprop optimizer to update the success discriminator. To make a fair comparison, our algorithm and R3L used the same exploration algorithm called RND \cite{burda2018exploration} as the reset policy. Table \ref{tab:hyperparameters} describes the key hyperparameters used in our experiments. We ran all experiments on a PC with a 3.20 GHz Intel i9-12900KF Processor, a GeForce RTX 2080 Ti GPU, and 64GB of RAM.

\subsection{Experimental Results and Analysis}
We use the following evaluation metrics: average episode step (AS), success rate (SR), and the average number of manual resets (MR). To compute the average episode step and the success rate, we sample initial states from a uniform distribution over the valid states. The average number of manual resets is recorded throughout the entire training procedure until performance converges. Note that we manually reset the environment when the agent fails to solve a task. The average episode step encodes how efficiently the agent solves a task and the success rate encodes the robustness of the agent’s performance. The average number of manual resets captures how many manual resets are required to train the agent to solve a task. 

Figure \ref{fig:quantitative_results_navigation} shows the learning curves computed over 10 random seeds for sparse-reward maze navigation tasks, and table \ref{tab:quantitative_results} describes their numerical training results computed over 100 episodes.
Reset RL obtains good performance in maze2d-1way-v1 and maze2d-2way-v1, but it fails to ensure robust performance in maze2d-4way-v1 and antmaze-4way-v2, where there are four paths to the goal. Oracle RL achieves higher success rates and lower average episode steps than Reset RL, as it has access to uniform initial state distributions. These results suggest that, even with access to extrinsic reset, resetting to diverse initial states is critical to achieving robust performance. Both Reset RL and Oracle RL rely on repetitive extrinsic resets after every episode, demanding substantial human intervention in the real world. In contrast, our algorithm consistently achieves robust performance in all tasks with significantly fewer manual resets than Reset RL and Oracle RL.

\begin{figure*}[t]
\vspace{0.1cm}
\begin{center}
\centerline{\includegraphics[width=0.60\textwidth, trim=8 8 8 8, clip]{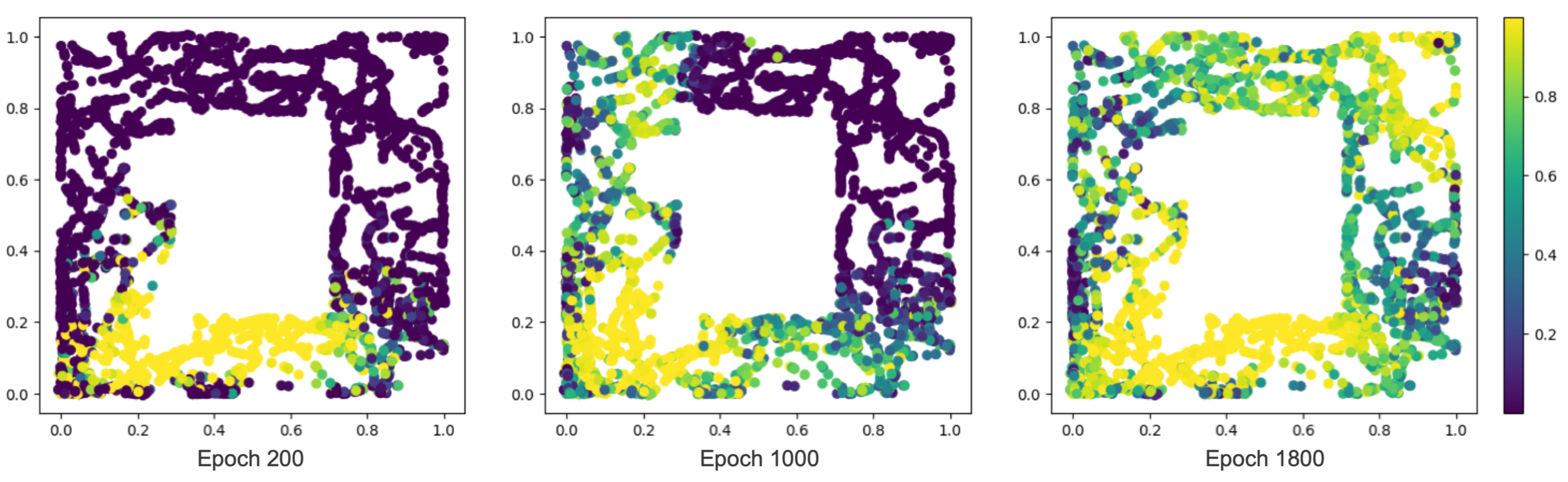}}
\vspace{-1.8mm}
\caption{Initial states allowed by our adaptive curriculum on maze2d-2way-v1. We normalize each dimension of states to [0,1] and their colors denote the success probabilities estimated by the success discriminator. Our adaptive curriculum allows initial states to be generated only near the goal at (0.1, 0.1) during the early stages of training. As training progresses, it also allows initial states to be generated from locations increasingly distant from the goal.}
\label{fig:init_states_2way}
\end{center}
\vskip -0.12in
\end{figure*}

\begin{figure*}[t]
\begin{center}
\centerline{\includegraphics[width=0.80\textwidth]{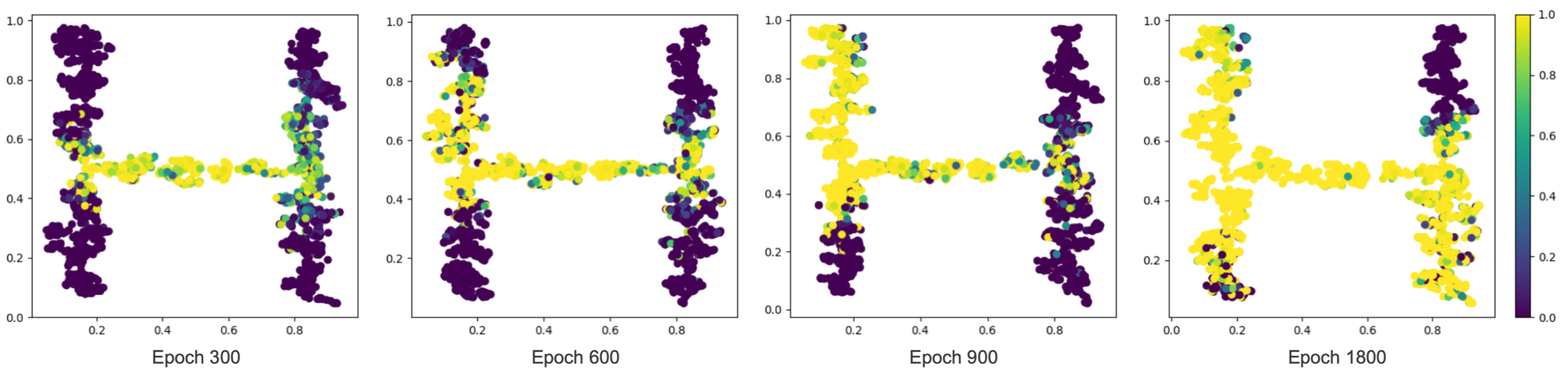}}
\vspace{-1.8mm}
\caption{Initial states allowed by our adaptive curriculum on antmaze-4way-v2. We normalize each dimension of states to [0,1] and their colors represent the success probabilities estimated by the success discriminator. Our adaptive curriculum generates diverse and informative initial states based on the learning progress of the forward policy, enabling the agent to perform efficient bootstrapping.}
\label{fig:init_states_4way}
\end{center}
\vskip -0.2in
\end{figure*}

It is unsurprising that LNT requires fewer manual resets than R3L and our algorithm, as LNT uses privileged information about the predefined initial state distribution. We also observed that the agent trained with LNT did not deviate from the initial states and failed to obtain informative transitions on maze2d-4way-v1 and antmaze-4way-v2. While this failure makes it easier for the agent to return to the initial state without manual resets, it also contributes to poor performance on these tasks. These results suggest that efficient bootstrapping is critical for ARL to ensure robust performance. Our algorithm achieves performance closest to Oracle RL, with a higher success rate and fewer manual resets than R3L on all navigation tasks. Furthermore, our algorithm is more stable and converges faster than R3L. These results imply that the capability of our curriculum to adapt to the learning progress of the forward policy can improve reset-free performance and sample efficiency. We would like to note that there is still room for improvement on antmaze-4way-v2. We leave further analysis of this room to future work, but we discuss several interesting directions to attain better performance in the next section.

\begin{figure}[t]
\begin{center}
\centerline{\includegraphics[width=0.93\columnwidth, trim=8 8 8 8, clip]{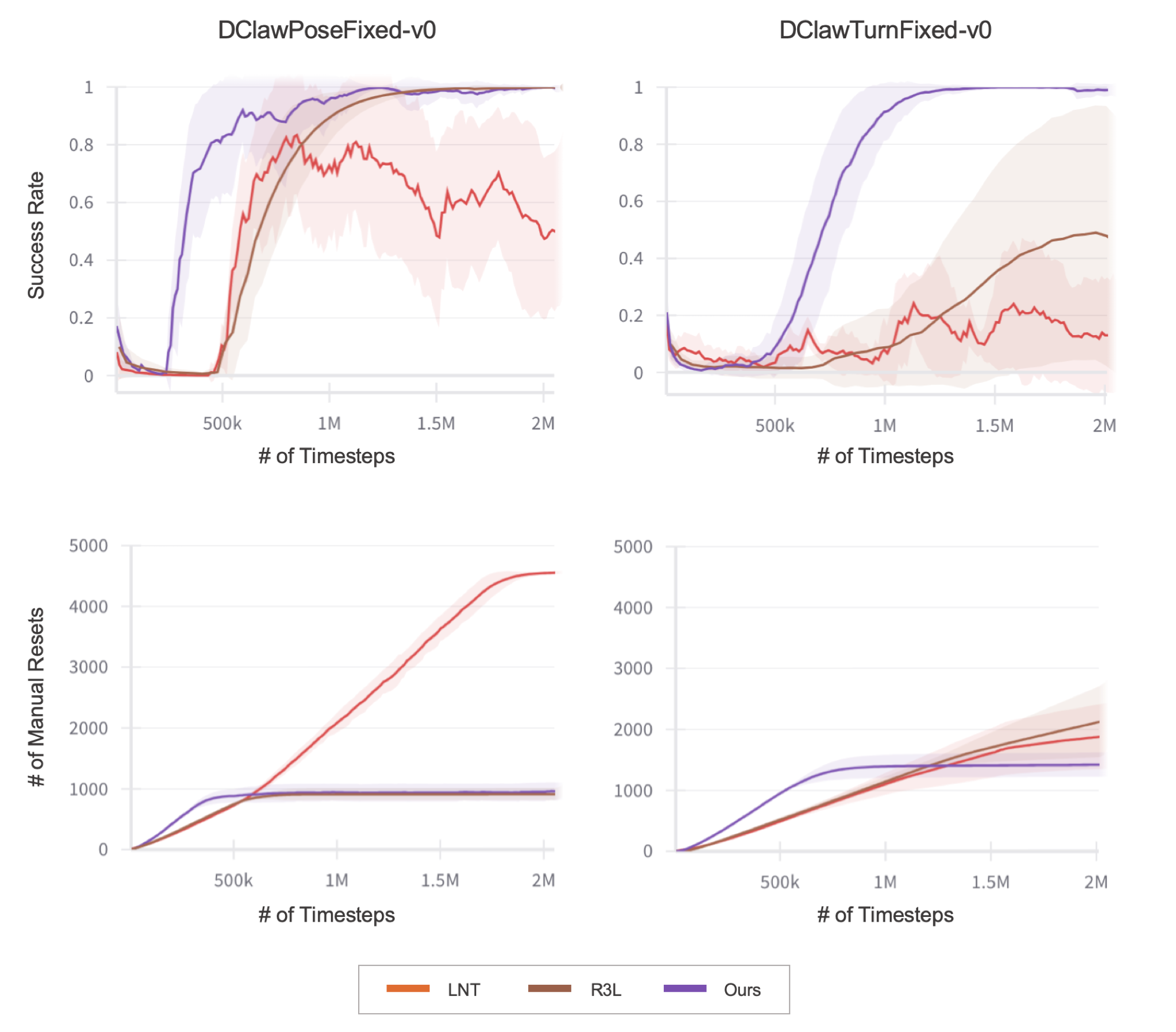}}
\vskip -0.1in
\caption{Learning curves for manipulation tasks. The darker-colored lines and shaded areas denote the means and standard deviations over 10 random seeds, respectively.}
\label{fig:quantitative_results_manipulation}
\end{center}
\vskip -0.3in
\end{figure}

Figure \ref{fig:quantitative_results_manipulation} describes the learning curves computed over 10 random seeds for sparse-reward manipulation tasks. We observed that LNT achieves lower success rates and requires more manual resets than R3L. This implies that in sparse reward tasks, where an agent can easily deviate from an initial state, LNT struggles to return the agent to the initial state, leading to unnecessary manual resets. Both R3L and our algorithm perform well on DClawPoseFixed-v0, requiring far fewer manual resets than LNT. However, unlike our algorithm, R3L suffers a significant performance decrease in DClawTurnFixed-v0. This suggests that the efficient bootstrapping enabled by our adaptive curriculum is crucial for tackling challenging tasks. We also would like to emphasize that our algorithm is task-agnostic so that it can be easily applied to a wide range of tasks beyond maze navigation and manipulation tasks.

We analyze how our algorithm creates an adaptive curriculum by visualizing the temporal changes in the initial states allowed by our algorithm. To do so, we sample states from rollout trajectories of the random policy that has access to uniform initial state distribution and use the success discriminator being trained to estimate the probability of solving a task for each sampled state, $C(s, a)$ where $a \sim \pi_r(a|s)$. Note that based on equation \ref{eq:initial_state_condition} and the hyperparameter table \ref{tab:hyperparameters}, the initial states allowed by our adaptive curriculum correspond to states with a success probability above 0.3 and below 0.7 when the agent follows the forward policy. 

Figure \ref{fig:init_states_2way} shows how the allowed initial states are changed over time on maze2d-2way-v1. We observed that our curriculum allows the agent to reset to initial states near the goal at the beginning of training and farther states away from the goal at the end of training. This suggests that our curriculum provides initial states adaptive to the learning progress of the forward policy and allows the agent to efficiently bootstrap on the success from easier initial states to solve a task from harder initial states. Figure \ref{fig:init_states_4way} describes how the allowed initial states are changed over time on antmaze-4way-v2. We observed that the initial states allowed by our curriculum exist only near the goal in the early stages of training but gradually spread out into different directions as the training progresses. This implies that our curriculum provides the agent with diverse and informative initial states, which enable it to achieve robust performance.

\begin{figure}[t]
\begin{center}
\centerline{\includegraphics[width=0.95\columnwidth, trim=8 8 8 8, clip]{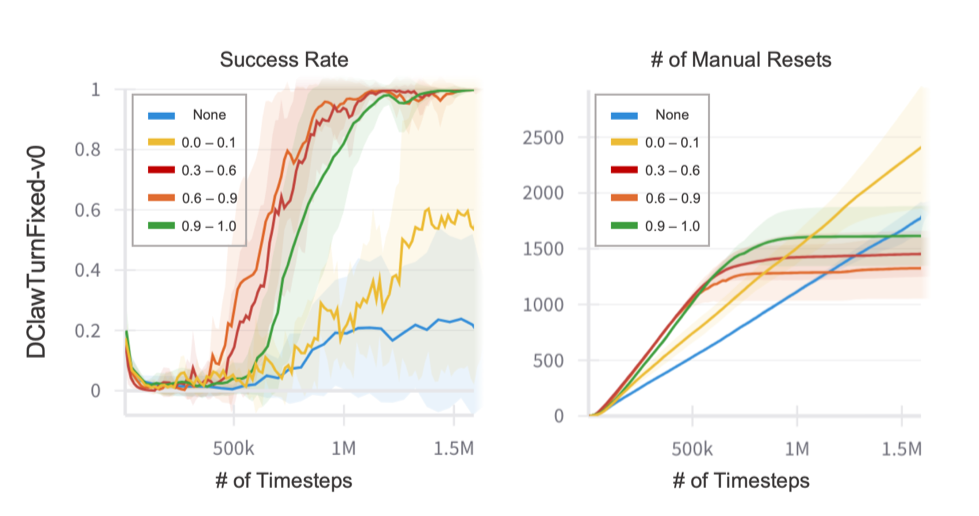}}
\vskip -0.05in
\caption{Ablation study on the success probability range of initial states. The darker-colored lines and shaded areas denote the means and standard deviations over 10 random seeds, respectively.}
\label{fig:ablation}
\end{center}
\vskip -0.25in
\end{figure}

Lastly, we conducted an ablation study to examine the effects of the success probability range of the initial states being sampled. The success probability range is determined by the two key hyperparameters, $\lambda_1$ and $\lambda_2$, and this range is the most critical design choice to generate our adaptive curriculum.  Figure \ref{fig:ablation} describes the performance according to the success probability range of the initial states on DClawTurnFixed-v0. We observed that excessively dangerous initial states ($\lambda_1=0.0$ and $\lambda_2=0.1$) cause more manual resets compared to random initial states (None). Initial states sampled with $\lambda_1=0.3$ and $\lambda_1=0.6$, or $\lambda_1=0.6$ and $\lambda_1=0.9$, allow an agent to obtain higher success rates and fewer manual resets than other initial states. This confirms the benefits of identifying diverse initial states that are informative and do not cause human intervention. It also indicates that our adaptive curriculum can identify such initial states without task-specific knowledge. Note that too safe initial states ($\lambda_1=0.9$ and $\lambda_2=1.0$) do not obtain the fewest manual resets, as an agent often fails to discover such initial states at the beginning of training.

\section{CONCLUSION}
We introduce a new ARL algorithm that generates an adaptive curriculum without task-specific knowledge. Our adaptive curriculum provides the agent with diverse and informative initial states, facilitating efficient bootstrapping and reducing manual resets. Experimental results demonstrate that our ARL algorithm enables the agent to solve sparse-reward maze navigation and manipulation tasks, outperforming baselines with much fewer manual resets. We will explore the following research directions in future work. First, we will investigate how to best integrate the benefits of goal-conditioned RL into our algorithm. We expect that this integration can scale our algorithm to contextual tasks. Second, we will explore optimization difficulties due to the non-stationary problem. We believe this research can improve the training stability of our success discriminator. Third, we will extend our algorithm to detect and avoid irreversible states, which are common in most real-world scenarios. This extension can allow our algorithm to be applied to more diverse real-world tasks. Finally, we will investigate whether training the reset policy to represent consistent and structured exploration behaviors can allow our algorithm to provide the agent with more diverse and informative initial states. We expect that leveraging unsupervised skill discovery algorithms, which discover diverse and distinct behaviors, is a promising approach for this research direction.


\bibliographystyle{./IEEEtran}
\bibliography{./IEEEabrv,./IEEEexample, ./references}

\end{document}